\documentclass[conference]{IEEEtran}

\usepackage{amsmath,amssymb,amsfonts}
\usepackage{graphicx}
\usepackage{textcomp}
\usepackage{booktabs}
\usepackage{multirow}
\usepackage{url}
\usepackage{siunitx}
\usepackage{algorithm}
\usepackage{algorithmic}
\usepackage{xcolor}
\usepackage{balance}
\usepackage{cite}
\usepackage{comment}
\usepackage{booktabs}
\usepackage{tabularx}
\usepackage{array}
\usepackage{makecell}
\usepackage[skip=2pt]{caption}

\usepackage{tikz}
\usetikzlibrary{shapes,arrows,positioning,calc,decorations.pathreplacing,fit,backgrounds,shadows,decorations.markings}

\usepackage{hyperref}
\begin{document}

\title{NanoBench: A Multi-Task Benchmark Dataset for\\Nano-Quadrotor System Identification,\\Control, and State Estimation}

\author{
\IEEEauthorblockN{Syed Izzat Ullah, Jos\'e Baca}
\IEEEauthorblockA{
Department of Engineering\\
Texas A\&M University-Corpus Christi, Corpus Christi, TX, USA\\
\{syed.izzatullah, jose.baca\}@tamucc.edu
}
}

\maketitle

\begin{abstract}
Existing aerial-robotics benchmarks target vehicles from hundreds of grams to several kilograms and typically expose only high-level state data. They omit the actuator-level signals required to study nano-scale quadrotors, where low-Reynolds-number aerodynamics, coreless DC motor nonlinearities, and severe computational constraints invalidate models and controllers developed for larger vehicles. We introduce \emph{NanoBench}, an open-source multi-task benchmark collected on the commercially available Crazyflie~2.1 nano-quadrotor (takeoff weight \SI{27}{\gram}) in a Vicon motion capture arena. The dataset comprises 172~flight recordings spanning multi-frequency excitation, geometric trajectory tracking at three speed regimes, and long-duration battery-drain hover. Each recording provides synchronized Vicon ground truth, raw IMU data, onboard extended Kalman filter (EKF) estimates, PID controller internals, and motor PWM commands at \SI{100}{\hertz}, alongside battery telemetry at \SI{10}{\hertz}, cross-aligned by gyroscope-based correlation on a \SI{1}{\milli\second} search grid. NanoBench defines standardized evaluation protocols, train/test splits, and open-source baselines for three tasks: nonlinear system identification, closed-loop controller benchmarking, and onboard state estimation assessment. To our knowledge, it is the first public dataset to jointly provide actuator commands, controller internals, and estimator outputs with millimeter-accurate ground truth on a commercially available nano-scale aerial platform.
\end{abstract}

\noindent 
\textbf{Anonymous dataset and code}:\\
\href{https://github.com/syediu/nanobench}{https://github.com/syediu/nanobench}\\
\textbf{Video}:
\href{https://syediu.github.io/assets/img/nanobench_demo.mp4}{https://syediu.github.io/assets/img/nanobench\_demo.mp4}

\section{Introduction}
\label{sec:introduction}

Learning-based methods have substantially advanced quadrotor autonomy. Deep reinforcement learning policies match professional pilots in drone racing~\cite{kaufmann2023champion}, adaptive controllers trained on minutes of flight data enable agile flight in strong wind~\cite{oconnell2022neural}, and model predictive controllers with learned residual dynamics outperform rigid-body baselines~\cite{torrente2021data}. These techniques depend on instrumented datasets for training and reproducible evaluation. Most publicly available datasets that support such evaluation were collected on platforms weighing hundreds of grams to several kilograms. For nano-scale quadrotors, platforms below \SI{50}{\gram}, that are widely used in embedded autonomy, safe learning, swarm robotics, autonomous navigation~\cite{aerocast, brunke2022safe, pichierri2023crazychoir, crazyswarm, ullah2025pofmader, syed}, no open benchmark exposes the full closed-loop stack.

Nano-scale quadrotors such as the Crazyflie~2.1 operate in aerodynamic and computational regimes that differ qualitatively from those of larger vehicles. Propellers with diameters below \SI{50}{\milli\meter} operate at Reynolds numbers on the order of $10^{4}$, where laminar separation bubbles and viscous losses significantly affect thrust and torque~\cite{deters2020reynolds}. Coreless DC motors introduce deadbands and response lags absent in brushless propulsion~\cite{forster2015system}, and proximity to surfaces induces strong ground-effect disturbances at low altitude~\cite{shi2019neural}. Computationally, a \SI{168}{\mega\hertz} Cortex-M4 microcontroller restricts onboard estimation to lightweight extended Kalman filters~\cite{palossi2019autonomous}.

These physical and computational constraints motivate systematic characterization of platform-level uncertainty in nano-scale aerial systems, including actuation nonlinearities, estimator behavior, and battery-dependent performance \cite{syedphddissertation}, yet no standardized benchmark exists to study nano-scale dynamics under these constraints. Thrust models transfer poorly across Reynolds-number regimes~\cite{bauersfeld2021neurobem}, policies trained in simulation degrade on physical hardware because of inaccurate thrust modeling~\cite{song2023reaching, eschmann2024learning}, and lightweight estimators require precisely synchronized multi-modal recordings for proper validation~\cite{elsheimy2020analysis}. Despite this demand, no existing dataset provides actuator commands, controller internals, and estimator outputs with external ground truth on a commercially available nano-scale platform. System identification, control, and state estimation methods for this class of vehicle are currently evaluated on private or synthetic data, which precludes systematic comparison~\cite{ullah2026syntrag}.

This paper presents \emph{NanoBench}, an open-source multi-task benchmark that addresses this gap. The dataset is collected on the standard Crazyflie~2.1 quadrotor inside a Vicon motion capture arena using a data collection framework built on direct radio communication through the cflib Python library. Our contributions are:

\begin{enumerate}
    \item \textbf{Nano-quadrotor flight dataset.} We release 172 flight recordings on the Crazyflie~2.1 covering multi-frequency excitation, geometric trajectory tracking at three speed regimes, and long-duration battery-drain hover. Every recording
    includes time-synchronized Vicon ground truth, raw IMU
    measurements, onboard EKF state estimates, PID controller
    internals, and motor PWM commands at \SI{100}{\hertz},
    with battery telemetry at \SI{10}{\hertz}.

    \item \textbf{Cross-correlation time alignment.} We develop
    and validate a synchronization procedure that cross-correlates
    onboard gyroscope rates with Vicon-derived angular
    velocity to estimate and correct the firmware-to-host clock
    offset. Across all flights, the residual misalignment is
    around \SI{1}{\milli\second}.

    \item \textbf{Multi-task evaluation suite.} We define
    train/test splits, metrics, and reporting conventions for
    three tasks (system identification, controller
    benchmarking, and state estimation) and release open-source
    baseline implementations that establish reference performance on each.
\end{enumerate}

The remainder of this paper is organized as follows. Section~\ref{sec:related_work} positions NanoBench against prior work. Section~\ref{sec:methodology} describes the platform, acquisition pipeline, synchronization procedure, trajectory design, and task formulations. Section~\ref{sec:experiments} reports dataset statistics and baseline results, Section~\ref{sec:limitations} provides the limitations, and Section~\ref{sec:conclusion} discusses the conclusions.

\begin{table*}[t]
\centering
\caption{Comparison of aerial robotics datasets. NanoBench is the first to combine actuator-level data, onboard estimator outputs, and controller internals on a commercially available nano-scale platform within a multi-task evaluation framework.}
\label{tab:dataset_comparison}

\resizebox{\textwidth}{!}{%
\begin{tabular}{@{}lccccc@{}}
\toprule
\textbf{Feature} &
\textbf{EuRoC}~\cite{burri2016euroc} &
\textbf{Blackbird}~\cite{antonini2020blackbird} &
\textbf{UZH-FPV}~\cite{delmerico2019uzh} &
\textbf{Busetto et al.}~\cite{busetto2025sysid} &
\textbf{NanoBench (Ours)} \\
\midrule
Platform mass & $\sim$\SI{2}{\kilo\gram} & $\sim$\SI{1}{\kilo\gram} & $\sim$\SI{0.8}{\kilo\gram} & $<$\SI{50}{\gram} & $\sim$\SI{27}{\gram} \\
Platform availability & Discontinued & Custom & Custom & Commercial & Commercial \\
Number of trajectories & 11 & 176 & 27 & 4 & 172 \\
Ground truth system & Vicon/Leica & OptiTrack & Leica & MoCap & Vicon \\
Ground truth rate & \SI{200}{\hertz} & \SI{360}{\hertz} & \SI{200}{\hertz} & Not specified & \SI{100}{\hertz} \\
Raw IMU data & \checkmark & \checkmark & \checkmark & -- & \checkmark \\
Motor commands & -- & \checkmark (RPM) & -- & \checkmark (RPM) & \checkmark (PWM) \\
Onboard estimator output & -- & -- & -- & -- & \checkmark \\
Controller internals & -- & -- & -- & -- & \checkmark \\
Battery telemetry & -- & -- & -- & -- & \checkmark \\
Camera data & Stereo & Synthetic & Stereo + event & -- & -- \\
Multi-task evaluation & -- & -- & -- & SysID only & SysID + Control + Estimation \\
Calibration tools & -- & -- & -- & -- & \checkmark \\
Open-source & \checkmark & \checkmark & Partial & \checkmark & \checkmark \\
\bottomrule
\end{tabular}%
}
\end{table*}


\section{Related Work}
\label{sec:related_work}

\subsection{Aerial Robot Datasets and Benchmarks}
\label{sec:rw_datasets}

Aerial robotics datasets can be partitioned by the downstream tasks they support and the platform scales they cover. The EuRoC MAV dataset~\cite{burri2016euroc}, UZH-FPV~\cite{delmerico2019uzh}, NTU VIRAL~\cite{nguyen2022ntu}, and INSANE~\cite{brommer2022insane} provide high-fidelity exteroceptive sensor data for visual-inertial odometry and multi-sensor fusion. However, these benchmarks treat the vehicle as a passive sensor carrier and omit actuator and control signals. As a result, they cannot support closed-loop dynamics modeling, controller evaluation, or estimator benchmarking.

Actuation-aware datasets capture the inputs necessary for system identification. The Blackbird dataset~\cite{antonini2020blackbird} and Agilicious framework~\cite{foehn2022agilicious} log motor commands for agile flight research. However, they feature quadrotors weighing near \SI{1}{\kilo\gram} equipped with brushless motors. These heavy platforms exhibit rigid-body dynamics that differ fundamentally from the coreless DC actuation and aerodynamic nonlinearities of nano-quadrotors, precluding direct transfer of models or control policies.

At the nano scale, Busetto et al.~\cite{busetto2025sysid} recently released a system identification dataset for a brushless Crazyflie variant that logs motor RPM. The dataset covers four trajectories and targets system identification only; it does not expose onboard estimator outputs, controller internals, or battery telemetry (Table~\ref{tab:dataset_comparison}). Multi-task evaluation on the standard, brushed Crazyflie~2.1 therefore remains unaddressed.

\subsection{System Identification for Aerial Vehicles}
\label{sec:rw_sysid}

Quadrotor dynamics modeling ranges from calibrated physics to fully learned representations. Forster~\cite{forster2015system} established baseline quadratic thrust parameters via least-squares fitting. Torrente et al.~\cite{torrente2021data} and NeuroBEM~\cite{bauersfeld2021neurobem} augment nominal rigid-body equations with data-driven residuals to capture unmodeled aerodynamic effects. Neural ordinary differential equations offer purely data-driven alternatives. Zhou et al.~\cite{zhuoNeuralODE} trained Neural ODEs for multi-step state prediction, while KNODE-MPC~\cite{chee2022knode} and O'Connell et al.~\cite{oconnell2022neural} embed adapted models within predictive controllers to improve tracking accuracy. 

Existing simulation ecosystems configure these learned and analytical models with static coefficients~\cite{llanes2024crazysim}. This constant-parameter assumption ignores the substantial thrust degradation caused by single-cell LiPo discharge, and no existing benchmark provides the synchronized voltage and kinematic data required to model this non-stationary effect.

\subsection{Control of Nano-Quadrotors}
\label{sec:rw_control}

The Crazyflie is a primary testbed for control architectures, including cascaded PID, geometric controllers on $SE(3)$, and minimum-snap trackers~\cite{mellinger2011minimum}. Safe reinforcement learning frameworks predominantly use it~\cite{brunke2022safe}, and recent work achieves zero-shot sim-to-real transfer~\cite{eschmann2024learning}. Song et al.~\cite{song2023reaching} showed that thrust calibration errors are the dominant source of tracking degradation during such transfer. Open-source ecosystems simplify swarm coordination and simulated execution~\cite{crazyswarm, pichierri2023crazychoir, llanes2024crazysim, cmas} and lower the barrier to multi-agent research.

Despite this active development, each research group runs its own private flight
experiments and reports numbers against its own reference
trajectories. Without a shared dataset, differences in results
between papers reflect hardware setup as much as algorithm
design.

\subsection{State Estimation for Micro Aerial Vehicles}
\label{sec:rw_estimation}

State estimation on sub-\SI{50}{\gram} platforms faces extreme hardware limitations. Visual-inertial odometry is too expensive for a \SI{168}{\mega\hertz}
Cortex-M4~\cite{palossi2022fully}, so nano-quadrotors run
lightweight EKFs with characterized IMU noise
models~\cite{elsheimy2020analysis, palossi2019autonomous}.

Evaluating these filters requires the estimator's
internal state recorded alongside an external
reference at matched timestamps; without that pairing, sensor noise cannot be separated from filter divergence.

\section{Methodology}
\label{sec:methodology}

\subsection{Hardware Platform}
\label{sec:hardware_platform}

We use the Crazyflie~2.1 nano-quadrotor (Bitcraze~AB) as the data collection platform. The vehicle measures $92 \times 92 \times
\SI{29}{\milli\meter}$ and weighs \SI{27}{\gram} in the stock configuration with a \SI{250}{\milli\ampere\hour} single-cell LiPo installed.
Four $7 \times \SI{16}{\milli\meter}$ coreless DC motors
drive \SI{46}{\milli\meter} four-blade propellers; at full
charge (\SI{4.2}{\volt}), combined peak thrust is
approximately \SI{0.6}{\newton}. All NanoBench recordings were flown in an instrumented configuration that adds retroreflective markers and the wireless-charging deck, raising the measured flying mass to \SI{40.85}{\gram}. The thrust-to-weight ratio is therefore $\approx\!2.2$ for the stock airframe but $\approx\!1.5$ as flown; we use the instrumented mass throughout. The onboard STM32F405 (\SI{168}{\mega\hertz} Cortex-M4F, 192\,kB SRAM) runs a cascaded PID controller and an extended Kalman filter; a secondary
nRF51822 handles radio communication. Inertial data are provided by a Bosch BMI088 six-axis IMU. Motor commands are 16-bit PWM values in $[0, 65535]$, and
the unregulated single-cell battery directly couples state of charge to actuator
performance.

Ground-truth 6-DoF pose is provided by a 12-camera Vicon system covering a
$6 \times 4 \times \SI{2}{\meter}$ volume. Retroreflective markers
(\SI{6.4}{\milli\meter} diameter) in an asymmetric configuration enable unique
rigid-body identification at \SI{100}{\hertz} with sub-millimeter accuracy.

\begin{table}[t]
\centering
\caption{Recorded signals and nominal sampling rates.}
\label{tab:signals}
\begin{tabular}{@{}llc@{}}
\toprule
\textbf{Signal} & \textbf{Source} & \textbf{Rate (Hz)} \\
\midrule
Position ($\mathbf{p} \in \mathbb{R}^3$), orientation ($\mathbf{q} \in S^3$) & Vicon & 100 \\
Linear velocity ($\dot{\mathbf{p}} \in \mathbb{R}^3$) & Vicon (derived) & 100 \\
Accelerometer ($\mathbf{a}_\text{IMU} \in \mathbb{R}^3$) [G] & BMI088 & 100 \\
Gyroscope ($\boldsymbol{\omega}_\text{IMU} \in \mathbb{R}^3$) [rad/s] & BMI088 & 100 \\
EKF state ($\hat{\mathbf{p}}, \hat{\dot{\mathbf{p}}}, \hat{\boldsymbol{\eta}}$) & Firmware & 100 \\
Controller setpoints ($\mathbf{p}_\text{ref}, \psi_\text{ref}$) & Firmware & 100 \\
PID outputs (roll, pitch, yaw, thrust) & Firmware & 100 \\
Motor PWM ($u_1, u_2, u_3, u_4 \in [0, 65535]$) & Firmware & 100 \\
Battery voltage $V_\text{bat}$ [V] & Firmware & 10 \\
\bottomrule
\end{tabular}
\end{table}

\subsection{Data Acquisition Pipeline}
\label{sec:pipeline}

\begin{figure}[t]
    \centering
    \includegraphics[width=1.0\columnwidth]{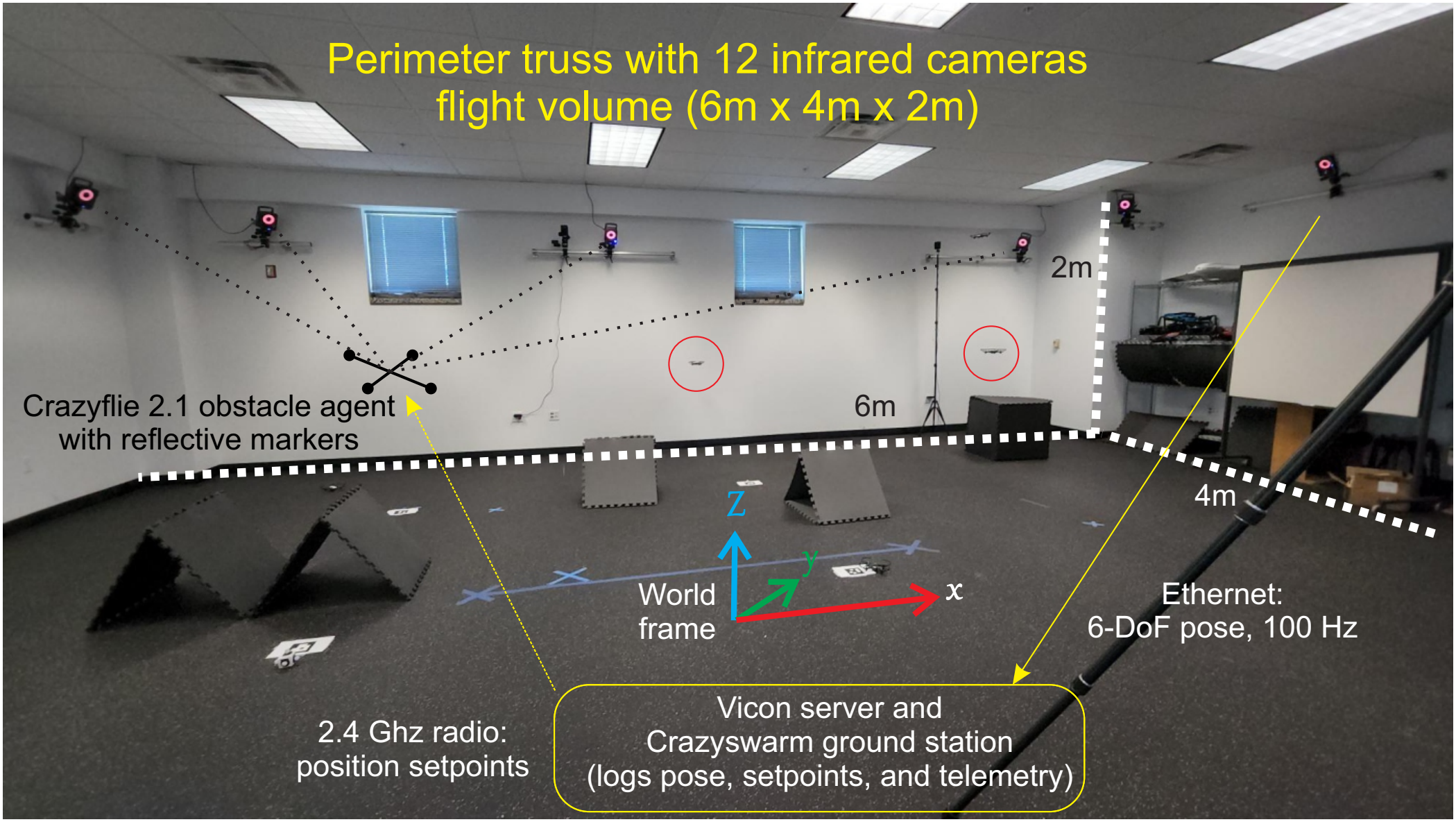}
    \caption{Experimental setup: a Crazyflie 2.1 inside the 12-camera Vicon arena, tracked at 100 Hz over a 6 × 4 × 2 m volume.}
    \label{fig:arena_annotated}
\end{figure}

Data collection uses a custom ROS 1 framework that communicates with the Crazyflie via cflib over a Crazyradio~PA USB dongle. The experimental setup is shown in Fig.~\ref{fig:arena_annotated}. Three concurrent data paths operate during each flight:

\textbf{Vicon path.} A \texttt{vrpn\_client\_ros} node publishes \texttt{geometry\_msgs/PoseStamped} messages on a per-rigid-body topic at \SI{100}{\hertz}. The \texttt{ViconRecorder} module subscribes to this topic, computes linear velocity by first-order backward difference, converts the quaternion to Euler angles, and writes each sample to a CSV file. On every callback, the pose is also forwarded to the Crazyflie EKF through cflib's \texttt{send\_extpose()} to provide the onboard estimator with an external position reference.

\textbf{Firmware telemetry path.} The \texttt{CflibLogger} module configures firmware log blocks from a YAML specification. Each block declares a set of firmware variables, their data types, and a polling frequency, which the cflib fetches over the CRTP radio protocol. On reception, each sample is stamped with both the host wall-clock time and the Crazyflie firmware tick (milliseconds since boot) and written to a per-block CSV file. Table~\ref{tab:signals} enumerates the recorded signals and their nominal polling rates. A single Crazyradio link is bandwidth-limited, so each block is delivered below its nominal rate and is resampled onto the uniform \SI{100}{\hertz} Vicon grid; raw per-block sample counts are released so the interpolated fraction of any signal is recoverable. Files carry both the Vicon quaternion and Euler angles in a fixed world frame.

\textbf{Command path.} The \texttt{ExperimentRunner} orchestrates the flight sequence. High-level commands (\texttt{takeoff}, \texttt{land}) use the firmware's onboard trajectory planner, which generates smooth polynomial setpoints internally. During trajectory execution, position setpoints are streamed at \SI{100}{\hertz} via \texttt{send\_position\_setpoint()}, which bypasses the high-level planner and feeds the position controller directly.

\begin{figure*}[htbp]
    \centering
    \includegraphics[width=\textwidth]{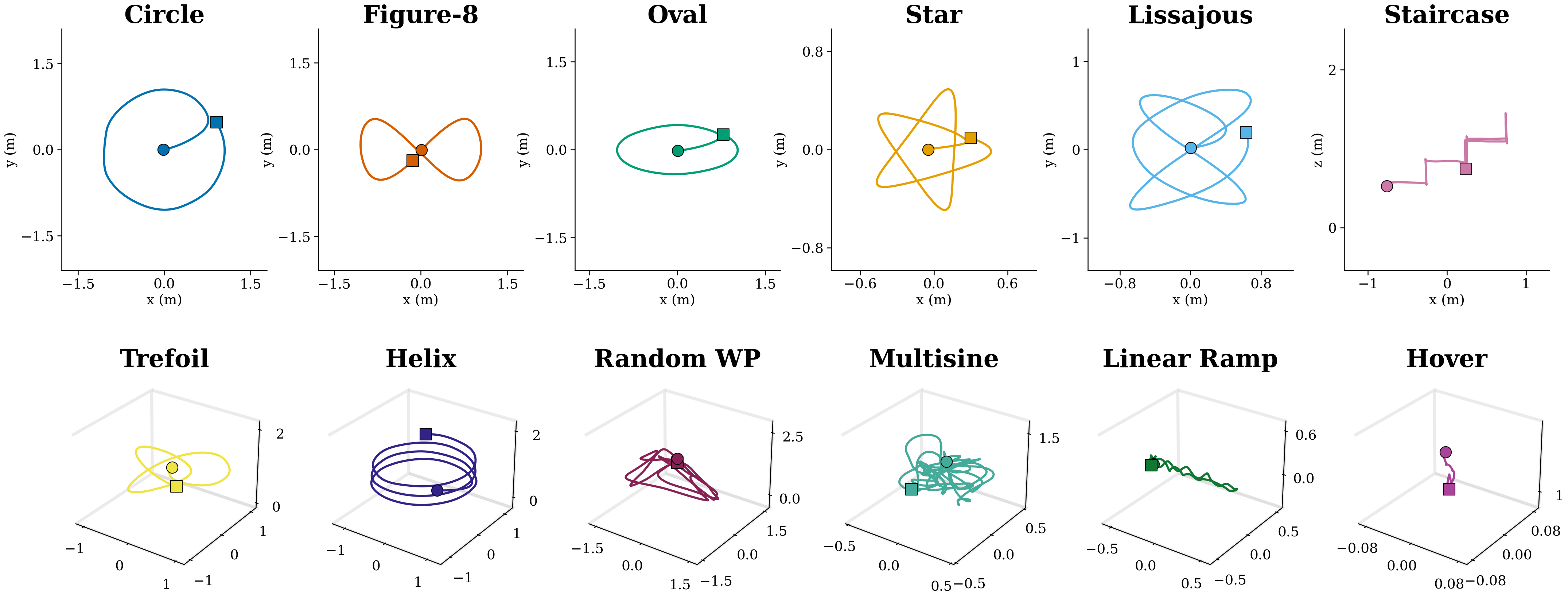}
    \caption{All 12 NanoBench trajectory types visualized from recorded Vicon ground-truth data. Categories A (excitation), B (geometric tracking), and C (battery-drain hover).}
    \label{fig:trajectory_design}
\end{figure*}
\subsection{Time Synchronization}
\label{sec:sync}

Three clocks run independently during each flight: the Vicon
Tracker, the Crazyflie firmware tick counter, and the ROS host.
The host timestamps Vicon poses and firmware packets on arrival,
which places both streams in a shared reference frame, but a residual offset remains from radio transmission latency (typically \SIrange{2}{5}{\milli\second}) and operating-system scheduling jitter.

We estimate this offset by cross-correlating the onboard
gyroscope rate $\boldsymbol{\omega}_\text{gyro}(t)$ with the angular
velocity $\boldsymbol{\omega}_\text{vicon}(t)$ obtained by
differentiating Vicon quaternions through a Savitzky--Golay
filter. Both signals are interpolated onto a common \SI{1}{\milli\second} grid and mean-removed, and the offset is
\begin{equation}
    \Delta t^* = \arg\max_{\Delta t \in [-2, 2]\,\text{s}} \; \sum_{k}
    \boldsymbol{\omega}_{\text{gyro}}(t_k) \cdot
    \boldsymbol{\omega}_{\text{vicon}}(t_k + \Delta t).
    \label{eq:xcorr}
\end{equation}
The sharp rotational features present during maneuvers localize the correlation peak to the \SI{1}{\milli\second} grid resolution, and the pipeline logs the peak correlation for each flight as an internal alignment-quality check. Once $\Delta t^*$ is found, firmware signals are shifted to the Vicon time base and resampled onto a uniform \SI{100}{\hertz} grid by linear interpolation. Each recording is then written to a single CSV file with all modalities on one time axis.

\subsection{Trajectory Design}
\label{sec:trajectories}

\begin{table}[t]
\centering
\caption{Category B: tracking trajectory parameters.}
\label{tab:trajectory_params}
\begin{tabular}{@{}llcc@{}}
\toprule
\textbf{Trajectory Type} & \textbf{Complexity Mode} & \textbf{Scale (m)} & \textbf{Speed (m/s)} \\
\midrule
Circle & Continuous yaw & $r=1.0$ & $0.5, 0.75, 1.0$ \\
Figure-8 & Direction reversal & $A \in \{1.0, 1.3\}$ & $0.5, 0.75, 1.0$ \\
Oval & Asymmetric axes & $1.0 \times 0.4$ & $0.5, 0.75, 1.0$ \\
Star & Sharp transients & $r \in \{0.6, 1.0\}$ & $0.5, 0.75, 1.0$ \\
Lissajous (3:2) & Dense planar & $A=0.6$ & $0.5, 0.75, 1.0$ \\
Trefoil Knot & 3D oscillation & $\Delta z=0.15$ & $0.5, 0.75, 1.0$ \\
Helix & Sweeping altitude & $\Delta z=1.0$ & $0.5, 0.75, 1.0$ \\
Linear Ramp & 1D acceleration & $L=2.0$ & $0.5, 0.75, 1.0$ \\
Random Waypoint & Unstructured 3D & Volume bound & $0.75$ \\
Staircase & Vertical steps & $\Delta z=0.5$ & $0.5$ \\
\bottomrule
\end{tabular}
\end{table}

The trajectory taxonomy comprises three categories and 12 trajectory types designed to jointly excite the platform's dynamic modes: system identification excitation, geometric tracking at multiple speeds, and long-duration battery-drain hover. Fig.~\ref{fig:trajectory_design} shows all 12 types from recorded flight data.

\textbf{Category A: system identification excitation.} Multi-sine trajectories superimpose sinusoidal position commands at logarithmically spaced frequencies from \SI{0.1}{\hertz} to \SI{5}{\hertz}:
\begin{equation}
    x(t) = \sum_{n=1}^{N_f} \frac{A}{n} \sin(2\pi f_n t), \quad
    y(t) = \sum_{n=1}^{N_f} \frac{A}{n} \cos(2\pi f_n t + \tfrac{\pi}{4}),
    \label{eq:multisine}
\end{equation}
where $A$ is the amplitude, $f_n$ are logarithmically spaced, and the phase offset between axes ensures simultaneous two-axis excitation. Category~A consists of a single trajectory type (\texttt{A1b\_multisine\_sysid}) executing a \SI{60}{\second} 3D multi-sine at an altitude of $\approx\SI{1.1}{\meter}$ with $N_f=15$ components in $[0.1, 5]\,\si{\hertz}$, amplitudes of \SI{0.8}{\meter} (horizontal) and \SI{0.35}{\meter} (vertical), and smooth \SI{2}{\second} fade-in and fade-out windows.

\textbf{Category B: geometric trajectory tracking.} Ten geometric trajectories exercise complementary dynamic modes at multiple speeds (Table~\ref{tab:trajectory_params}). Most shapes are parameterized symmetrically, e.g., circular paths $x(t) = r\cos(\omega t)$, $y(t) = r\sin(\omega t)$ with $\omega = v / r$. Execution speeds range from \SIrange{0.5}{1.0}{\meter\per\second}, the practical operating envelope of nano-quadrotors in confined environments.

\textbf{Category C: battery drain.} Four long-duration hover flights (\texttt{C4\_battery\_drain}) provide continuous recordings from $\approx\SI{4.2}{\volt}$ down to $\approx\SI{3.1}{\volt}$ while the vehicle holds position at $\approx\SI{1.0}{\meter}$, capturing voltage-related thrust degradation across the full discharge curve. These flights last 101--\SI{134}{\second} and support voltage-conditioned thrust modeling and low-voltage evaluation conditions.

\subsection{Benchmark Task Formulation}
\label{sec:benchmark}

NanoBench establishes standardized protocols for three evaluation tasks.

\textbf{Task 1: system identification.} Given the initial state and the sequence of motor commands $\{u_i^{(k)}\}$, a model must predict the 6-DoF state trajectory $\hat{\mathbf{x}}$ over open-loop horizons of up to $H=50$ steps (\SI{500}{\milli\second} at \SI{100}{\hertz}); results are reported at $h \in \{1, 10, 50\}$. The reference formulation is rigid-body dynamics in the world frame $\mathcal{W}$:
\begin{align}
    m \ddot{\mathbf{p}} &= \mathbf{R}(\mathbf{q}) \begin{bmatrix} 0 \\ 0 \\
    \textstyle\sum_i T_i \end{bmatrix} - mg\,\mathbf{e}_3 - \mathbf{D}_t \dot{\mathbf{p}},
    \label{eq:trans_dyn} \\[4pt]
    \mathbf{J}\,\dot{\boldsymbol{\omega}} &= \boldsymbol{\tau} -
    \boldsymbol{\omega} \times \mathbf{J}\boldsymbol{\omega} - \mathbf{D}_r \boldsymbol{\omega},
    \label{eq:rot_dyn}
\end{align}
where $\mathbf{R}(\mathbf{q}) \in SO(3)$ is the body-to-world rotation, $\mathbf{D}_t$ and $\mathbf{D}_r$ are diagonal drag matrices, and $\boldsymbol{\tau}$ is the torque vector. Training uses the Category~A excitation flights at nominal voltage ($V > \SI{3.8}{\volt}$); testing uses held-out Category~B tracking flights.

For each rollout initialized at $t \in \mathcal{T}$, the model generates open-loop predictions $\hat{x}_{t+h|t}$ for $h = 1,\dots,H$. The component-wise prediction error at horizon $h$ is
\begin{equation}
e^{(c)}_{t,h} =
\begin{cases}
\left\| x^{(c)}_{t+h} - \hat{x}^{(c)}_{t+h|t} \right\|_2, & c \in \{p, v, \omega\}, \\[6pt]
2\,\mathrm{atan2}\!\left(\left\|q_v\right\|_2,\, q_w\right), & c = R,
\end{cases}
\label{eq:error_def}
\end{equation}
where $x^{(c)}$ denotes the corresponding Euclidean state component and $[q_v \; q_w]^\top = q_{t+h}^{-1} \otimes \hat{q}_{t+h|t}$ is the relative quaternion. The mean absolute error at horizon $h$ and its cumulative $H$-step variant are
\begin{equation}
\mathrm{MAE}^{(c)}(h) = \frac{1}{|\mathcal{T}|} \sum_{t \in \mathcal{T}} e^{(c)}_{t,h},
\qquad
\mathrm{MAE}^{(c)}_{1:H} = \sum_{h=1}^{H} \mathrm{MAE}^{(c)}(h).
\label{eq:mae_def}
\end{equation}

\textbf{Task 2: controller benchmarking.} This task evaluates closed-loop tracking of Category~B reference trajectories $\mathbf{p}_{\text{ref}}(t)$. Evaluation flights are matched on initial battery voltage ($\Delta V \leq \SI{0.05}{\volt}$) to isolate controller performance from voltage-induced thrust variation. The position tracking error at time step $k$ is
\begin{equation}
    e_p(k) = \bigl\| \mathbf{p}(t_k) - \mathbf{p}_{\text{ref}}(t_k) \bigr\|_2 .
    \label{eq:tracking_error}
\end{equation}
Aggregate performance is reported as position RMSE, average displacement error (ADE, the mean of $e_p$), velocity RMSE, mean heading error $\bar{e}_\psi$, and a divergence rate defined as the percentage of time steps with $e_p(k) > \SI{0.5}{\meter}$. Real-flight controllers are evaluated over repeated flights and reported as mean~$\pm$~standard deviation; offline controllers use deterministic rollouts and are reported without repetition.

\textbf{Task 3: state estimation.} This task isolates the accuracy of lightweight onboard estimators against Vicon ground truth. The estimator is configured at takeoff and runs for the duration of that flight. Estimator output $\hat{\mathbf{p}}(t_k)$ is evaluated against ground truth $\mathbf{p}_{\text{gt}}(t_k)$ using the absolute trajectory error (ATE) after rigid alignment by Horn's $SE(3)$ method:
\begin{equation}
    \mathrm{ATE} = \left( \frac{1}{N} \sum_{k=1}^{N}
    \bigl\| \mathbf{p}_{\text{gt}}(t_k) - \mathbf{T}\,\hat{\mathbf{p}}(t_k)
    \bigr\|_2^2 \right)^{1/2},
    \label{eq:ate}
\end{equation}
where $\mathbf{T}$ is the rigid transformation returned by the alignment. Additional metrics are the relative trajectory error (RTE) over sliding path-length windows, per-axis velocity residuals, and attitude RMSE in Euler angles. All flights of a trajectory type are pooled per speed regime and reported as mean~$\pm$~standard deviation across runs, with the number of runs $N$ stated per condition.

\section{Experimental Validation and Results}
\label{sec:experiments}

\subsection{Dataset Statistics}
\label{sec:dataset_stats}

\texttt{NanoBench} contains 172 flight recordings spanning the 12 trajectory types of Categories~A--C, with a total aligned flight time of \SI{107}{\minute} and 641{,}915 synchronized samples at \SI{100}{\hertz}. Table~\ref{tab:trajectory_summary} reports per-category statistics.

\begin{table}[t]
\centering
\caption{Trajectory categories in the NanoBench dataset.}
\label{tab:trajectory_summary}
\small
\setlength{\tabcolsep}{3pt}

\begin{tabularx}{\columnwidth}{
    @{}
    l
    >{\raggedright\arraybackslash}X
    c
    c
    c
    @{}
}
\toprule
\textbf{Category} &
\textbf{Trajectory types} &
\textbf{Flights} &
\makecell{\textbf{Flight}\\\textbf{time}} &
\textbf{Samples} \\
\midrule

A: Excitation &
Multi-sine (3D) &
4 &
\SI{4.6}{\minute} &
27,518 \\

B: Tracking &
Circle, figure-8, oval, star, trefoil, Lissajous,
helix, ramp, random waypoints, staircase &
164 &
\SI{96.4}{\minute} &
578,623 \\

C: Battery drain &
Battery-drain hover &
4 &
\SI{6.0}{\minute} &
35,774 \\

\midrule

\textbf{Total} &
&
\textbf{172} &
\textbf{\SI{107.0}{\minute}} &
\textbf{641,915} \\

\bottomrule
\end{tabularx}
\end{table}

Battery voltage across the dataset spans \SIrange{3.08}{4.2}{\volt}, which covers the full operational discharge curve of a single-cell LiPo. Category~B tracking flights operate in the mid-to-nominal voltage regime ($\SIrange{3.8}{4.2}{\volt}$), while Category~C hovers capture the complete discharge from $\approx\SI{4.2}{\volt}$ to $\approx\SI{3.1}{\volt}$.

The synchronization pipeline (Section~\ref{sec:sync}) was applied to all 172 recordings. The estimated clock offset has a median of \SI{282}{\milli\second} with a standard deviation of \SI{17.3}{\milli\second} across flights. This offset is dominated by the difference between host wall-clock initialization and the firmware tick origin; radio transmission contributes only \SIrange{2}{5}{\milli\second} of additional jitter, which is reflected in the standard deviation. After correction, the residual alignment error is bounded by the \SI{1}{\milli\second} search grid.

\subsection{Task 1: System Identification Baselines}
\label{sec:sysid_results}

\subsubsection{Baselines}
\label{sec:sysid_baselines}

We evaluate five baseline models of increasing complexity, adapted from Busetto et al.~\cite{busetto2025sysid}. All learned models are trained on the multi-sine excitation flights using a temporal 80/20 split, in which the first \SI{80}{\percent} of each flight forms the training set and the final \SI{20}{\percent} the validation set, so that no prediction window spans both. Testing uses held-out Category~B tracking flights (Table~\ref{tab:trajectory_params}) that contribute no training data. Below, we briefly explain the baselines for Task 1.

\textbf{Naive model.} A constant extrapolation model that predicts $\hat{\mathbf{x}}(t_k + h) = \mathbf{x}(t_k)$ for all horizons $h$, which serves as a lower bound.

\textbf{Physics model.} A rigid-body model implementing~\eqref{eq:trans_dyn}--\eqref{eq:rot_dyn} with a quadratic motor-thrust map and fourth-order Runge--Kutta integration at \SI{100}{\hertz}. The bare airframe weighs \SI{27}{\gram}; with three retroreflective markers and the wireless-charging deck attached, the measured flying mass is \SI{40.85}{\gram}, which all dynamics models use. Inertia and drag parameters are taken from~\cite{busetto2025sysid}. Because no torque model is fitted, angular acceleration is held at zero ($\dot{\boldsymbol{\omega}} \equiv 0$), so attitude is propagated from the initial angular rate. This model is not learned; it uses fixed, calibrated parameters.

\begin{figure}[t]
    \centering
    \includegraphics[width=1.0\columnwidth]{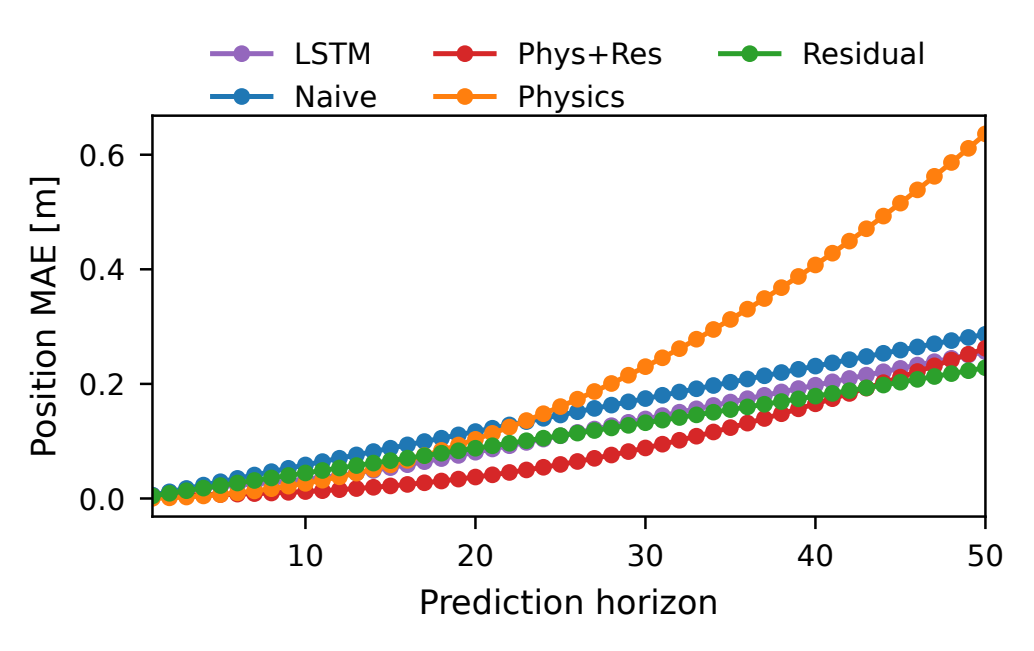}
    \caption{Task 1: position MAE vs. prediction horizon. The physics model is most accurate at $h=1$ but diverges beyond roughly $h=15$ steps; the physics+residual hybrid model achieves the lowest cumulative error.}
    \label{fig:task1_pos_mae_horizon}
\end{figure}

\textbf{Residual MLP.} A feedforward network operating in the 12-D state space (position, velocity, $SO(3)$ rotation vector, angular velocity) predicts a residual $\Delta \mathbf{x}$ on top of the current state. The network consists of five fully connected layers with 64 hidden units and ReLU activations. It is trained to minimize a horizon-weighted mean-squared error over 50-step prediction windows using Adam with a cosine-annealed learning rate.

\textbf{Physics + residual.} A hybrid model that composes the physics baseline with the residual MLP. The physics model is rolled out to obtain a normalized prediction $\hat{\mathbf{x}}_{\text{phys}}$, and the residual network predicts a correction in the same normalized space.

\textbf{LSTM.} A recurrent model conditioned on the initial state $\mathbf{x}_0$ and the sequence of motor commands. A single-layer LSTM with 64 hidden units processes the control sequence and outputs per-step state increments that are accumulated into predicted trajectory. As in the MLP case, training minimizes a horizon-weighted MSE over 50-step windows.

For all learned models we use batch size~256, an exponential temporal weighting factor $\lambda = 0.1$ in the loss, and train for 500~epochs.

\subsubsection{Results and Analysis}

    \begin{table*}[t]
        \centering
        \caption{Task 1: MAE per state component at horizons $h=1,10,50$ steps. The italic column reports the cumulative error (sum of MAEs over $h=1,\dots,50$). Best value per column in bold.}
        \setlength{\tabcolsep}{3pt}
        \scriptsize
        \renewcommand{\arraystretch}{1.2}
        \begin{tabular}{l|cccc|cccc|cccc|cccc}
            \toprule
            & \multicolumn{4}{c|}{$\mathrm{MAE}_{p,h}$ [m]}
            & \multicolumn{4}{c|}{$\mathrm{MAE}_{v,h}$ [m/s]}
            & \multicolumn{4}{c|}{$\mathrm{MAE}_{R,h}$ [rad]}
            & \multicolumn{4}{c}{$\mathrm{MAE}_{\omega,h}$ [rad/s]}\\[1mm]

            Model
            & $h{=}1$ & $h{=}10$ & $h{=}50$ & \textit{$h{=}1{:}50$}
            & $h{=}1$ & $h{=}10$ & $h{=}50$ & \textit{$h{=}1{:}50$}
            & $h{=}1$ & $h{=}10$ & $h{=}50$ & \textit{$h{=}1{:}50$}
            & $h{=}1$ & $h{=}10$ & $h{=}50$ & \textit{$h{=}1{:}50$}\\
            \midrule
Naive & 0.0057 & 0.0585 & 0.2866 & 7.3866 & \textbf{0.0085} & \textbf{0.0786} & \textbf{0.3542} & \textbf{9.4178} & 0.0033 & 0.0285 & \textbf{0.0915} & \textbf{2.7563} & \textbf{0.0402} & \textbf{0.2602} & \textbf{0.4524} & \textbf{17.1131} \\
Physics & \textbf{0.0003} & 0.0265 & 0.6364 & 10.9641 & 0.0511 & 0.5085 & 2.5673 & 65.0322 & \textbf{0.0019} & \textbf{0.0198} & 0.1425 & 3.2693 & \textbf{0.0402} & \textbf{0.2602} & \textbf{0.4524} & \textbf{17.1131} \\
Residual & 0.0045 & 0.0445 & \textbf{0.2282} & 5.6810 & 0.0121 & 0.1096 & 0.6708 & 16.9002 & 0.0084 & 0.1051 & 0.1971 & 8.8737 & 0.2043 & 1.1173 & 1.3046 & 54.1783 \\
Phys+Res & 0.0036 & \textbf{0.0121} & 0.2621 & \textbf{4.4071} & 0.0201 & 0.1370 & 1.0771 & 26.7469 & 0.0065 & 0.1359 & 0.1825 & 9.8645 & 0.8784 & 1.5799 & 1.1781 & 55.3050 \\
LSTM & 0.0058 & 0.0332 & 0.2562 & 5.9194 & 0.0199 & 0.1077 & 0.4782 & 11.9639 & 0.0162 & 0.1339 & 0.3059 & 9.8302 & 0.1256 & 0.8159 & 5.1525 & 118.0662 \\
            \bottomrule
        \end{tabular}
        \label{tab:numerical_performance}
    \end{table*}

Table~\ref{tab:numerical_performance} reports MAE per state component at $h \in \{1, 10, 50\}$ steps and the cumulative score over $h = 1,\dots,50$ steps and the cumulative score over $h = 1\ldots50$.
Based on Table~\ref{tab:numerical_performance} and Fig.~\ref{fig:task1_pos_mae_horizon}, we find the following three observations:

\textbf{First-principles accuracy degrades rapidly beyond short horizons.}
At $h{=}1$ (\SI{10}{\milli\second}), the physics model achieves sub-millimeter position
MAE (\SI{0.3}{\milli\meter}), significantly lower than naive extrapolation
(\SI{5.7}{\milli\meter}), which shows that rigid-body dynamics accurately describe the
immediate state evolution of the platform. However, the advantage reverses sharply at longer
horizons, by $h{=}50$, physics-model position MAE reaches
\SI{636}{\milli\meter}, 2.2 times worse than naive. The
velocity channel reveals the mechanism: physics-model velocity MAE grows from
\SI{51}{\milli\meter\per\second} at $h{=}1$ to \SI{2.57}{\meter\per\second} at $h{=}50$,
which indicates that small thrust-modeling errors compound through integration
and corrupt the translational state. 

\textbf{The hybrid architecture achieves the best cumulative accuracy.}
The physics + residual model attains the lowest cumulative position error (\SI{4.41}{\meter}), ahead of the residual MLP (\SI{5.68}{\meter}) and the LSTM (\SI{5.92}{\meter}). The advantage is most distinct
at $h{=}10$, where it reaches \SI{12.1}{\milli\meter}
position MAE, 3.7 times lower than the residual MLP and 2.7 times lower than the LSTM.
At $h{=}50$, however, the pure residual MLP achieves the lowest position MAE
(\SI{228}{\milli\meter}), which suggests that the physics component eventually introduces
compounding errors that the residual correction only partially absorbs. The crossover
between these two models near $h = 30$ defines a practical horizon boundary for each
architecture.

\begin{figure}[t]
    \centering
    \includegraphics[width=1.0\columnwidth]{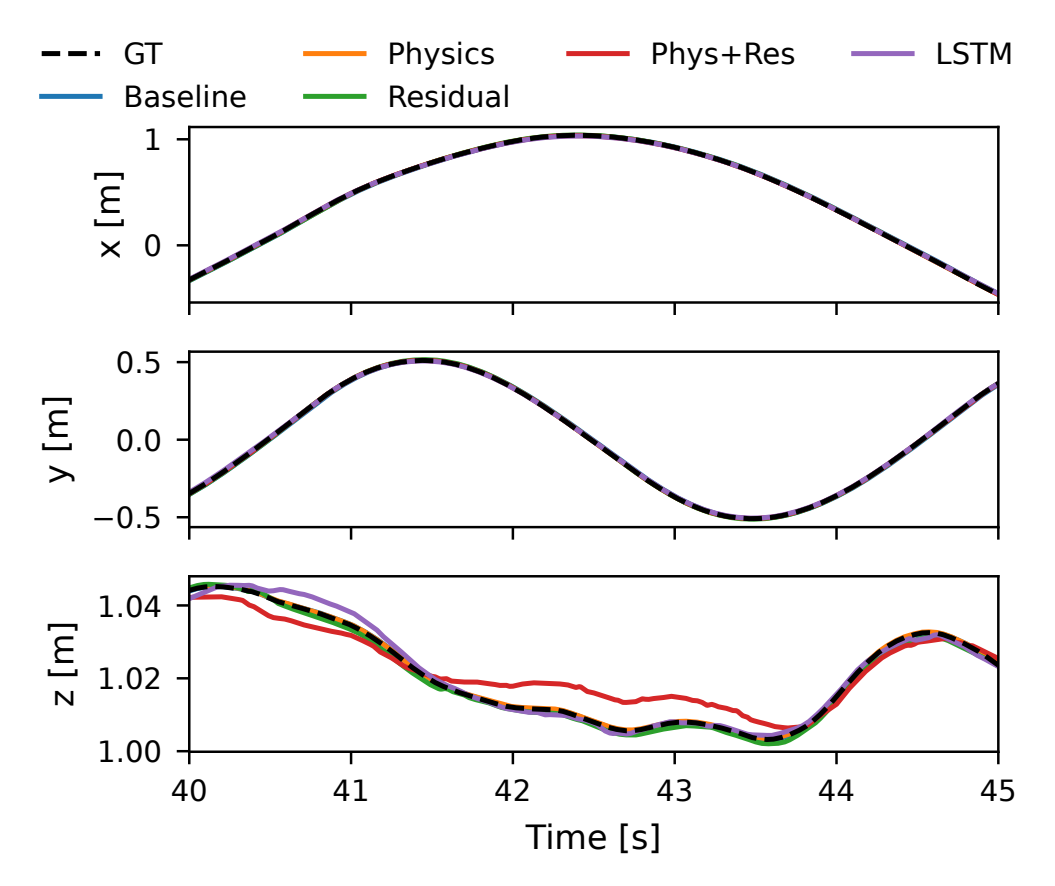}
    \caption{Task 1: predicted (colored) versus Vicon ground-truth (black) trajectories for each baseline on a held-out Category~B helix sequence, per axis.}
    \label{fig:task1_baselines_vs_gt}
\end{figure}

\textbf{Translational and rotational objectives trade off under a joint loss.}
The LSTM achieves competitive velocity MAE (\SI{0.48}{\meter\per\second} at $h{=}50$,
second only to naive), but its angular-velocity MAE reaches \SI{5.15}{\radian\per\second},
an order of magnitude above the naive baseline (\SI{0.45}{\radian\per\second}).
The physics model produces identical angular velocity predictions to naive at all
horizons because the implementation holds $\dot{\boldsymbol{\omega}} \equiv 0$ (Section~\ref{sec:sysid_baselines}),
which reflects the difficulty of modeling torque transients from coreless DC motors
without dedicated angular acceleration data. Fig.~\ref{fig:task1_baselines_vs_gt} visualizes per-axis predictions against ground truth on a held-out helix as a test sequence.

\subsection{Task 2: Controller Benchmarking}
\label{sec:results_control}

\subsubsection{Baselines}
\label{sec:control_baselines}
We benchmark five controllers under two evaluation paradigms that separate physical hardware performance from data-driven rollout. \textbf{Real-flight} controllers execute onboard the platform with Vicon-aided state estimation: (1)~\textbf{PID}, the firmware's default cascaded PID controller; and (2)~\textbf{Mellinger}, a geometric tracking controller on $SE(3)$~\cite{mellinger2011minimum}. \textbf{Offline-learned} controllers are trained from logged flight data and evaluated by closed-loop rollouts against a learned dynamics model: (3)~\textbf{BC-MLP} and (4)~\textbf{BC-LSTM}, behavior-cloning policies trained by supervised imitation of the onboard PID using the \texttt{imitation} framework~\cite{gleave2022imitation}; and (5)~\textbf{MPPI}~\cite{williams2017information}, a model predictive path integral controller using a neural forward model trained on the same data, run with a 10-step (\SI{100}{\milli\second}) horizon, 128 sampled rollouts, temperature $\lambda = 0.1$, and exploration noise $\sigma = 0.02$ on normalized commands. Because the physical environment and the learned simulator differ fundamentally, hardware and offline metrics are reported separately in Table~\ref{tab:controller_benchmarking} and are not directly comparable.

\subsubsection{Results and Analysis}
\label{sec:control_results}
\begin{table}[t]
\caption{Task 2: controller benchmarking results. Real-flight rows report mean~$\pm$~standard deviation over repeated flights; offline rows are deterministic rollouts on the learned plant and carry no spread.}
\label{tab:controller_benchmarking}
\centering
\resizebox{\columnwidth}{!}{
\begin{tabular}{llccccc}
\toprule
Type & Method & RMSE$_p$$\downarrow$ & ADE$_p$$\downarrow$ & RMSE$_v$$\downarrow$ & $\bar{e}_\psi$$\downarrow$ & Div.\,(\%)$\downarrow$ \\
     &        & (m) & (m) & (m/s) & (deg) & \\
\midrule
\multirow{2}{*}{Real flight}
 & PID       & 0.29$\pm$0.03 & 0.27$\pm$0.02 & 0.72$\pm$0.14 & 5.82$\pm$0.58 & 4.20$\pm$2.46 \\
 & Mellinger & 0.28$\pm$0.00 & 0.27$\pm$0.00 & 0.89$\pm$0.18 & 5.27$\pm$1.61 & 0.01$\pm$0.01 \\
\midrule
\multirow{3}{*}{Offline learned}
 & BC-MLP    & 0.15 & 0.13 & 0.19 & 1.07 & 0.00 \\
 & BC-LSTM   & 0.18 & 0.16 & 0.22 & 0.97 & 0.00 \\
 & MPPI      & 1.32 & 1.13 & 0.72 & 4.50 & 75.22 \\
\bottomrule
\end{tabular}
}
\end{table}

\begin{figure}[htbp]
    \centering
    \includegraphics[width=1.0\columnwidth]{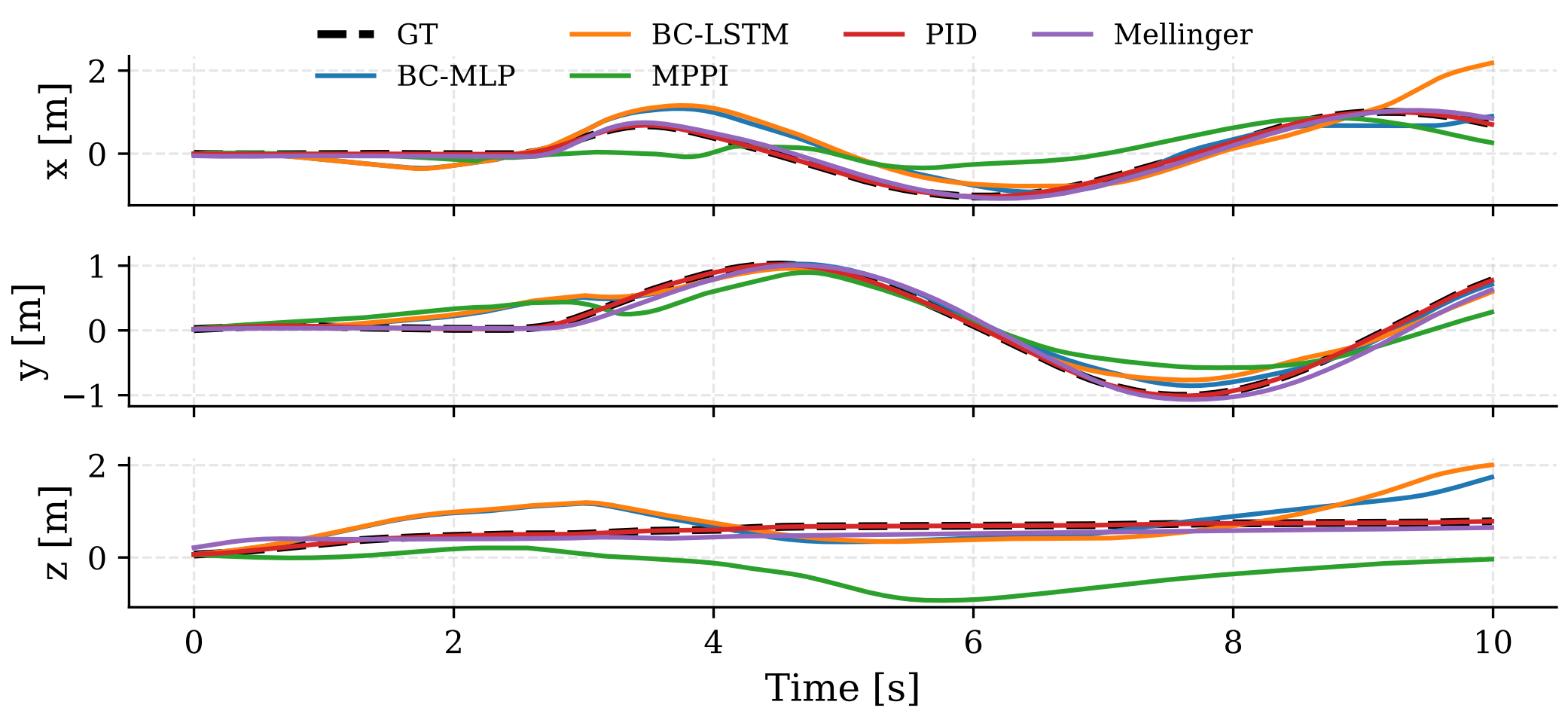}
    \caption{Task 2: closed-loop position (m) per axis for each baseline. The dashed trace is the commanded reference trajectory, not a measured vehicle state. Real-flight (PID, Mellinger) and offline (BC-MLP, BC-LSTM, MPPI) trajectories.}
    \label{fig:task2_baselines_vs_gt}
\end{figure}

\begin{figure}[htbp]
    \centering
    \includegraphics[width=\columnwidth]{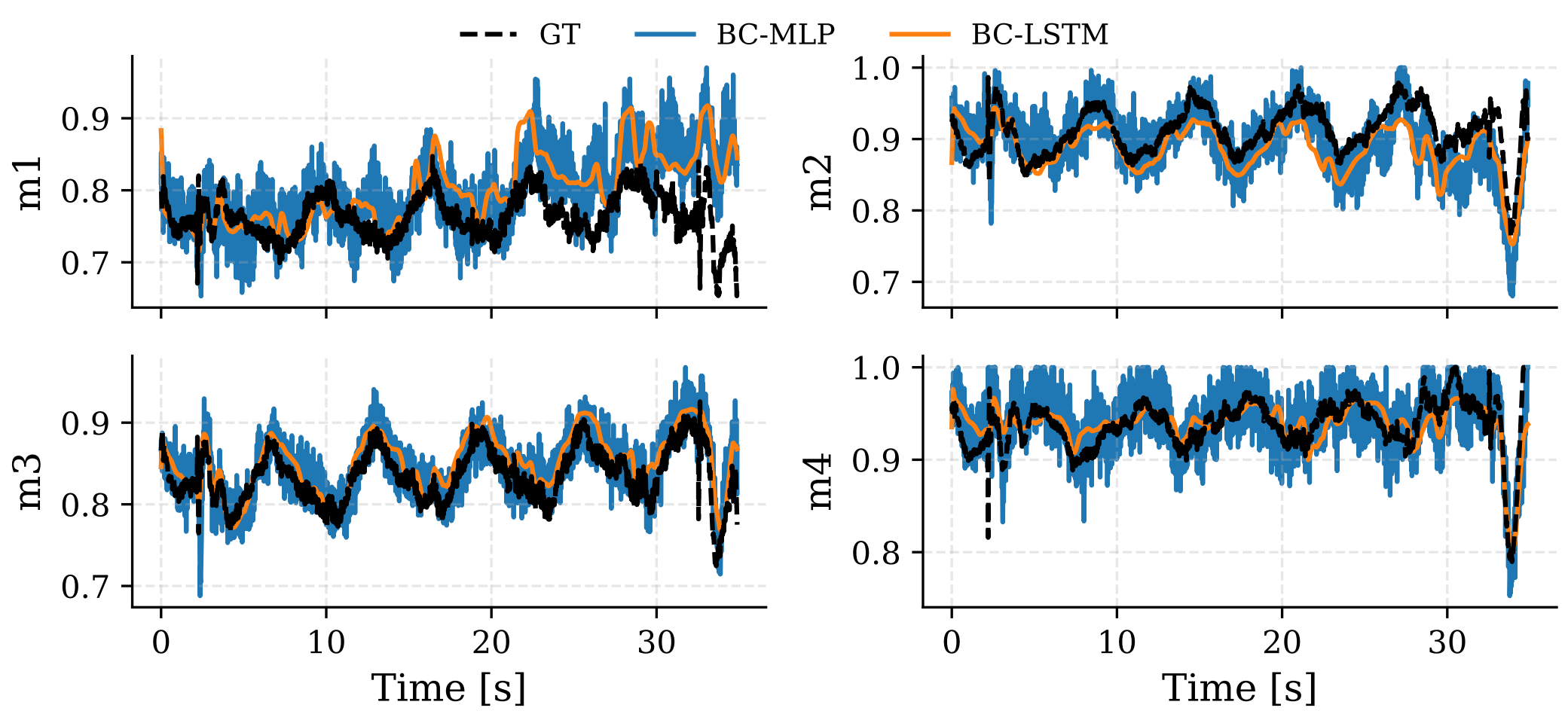}
    \caption{Task 2: normalized motor PWM commands over time for the two behavior-cloning policies against the logged expert commands (dashed). The real-flight controllers are omitted, as their commands are not generated on the learned plant.}
    \label{fig:task2_baselines_motors_overtime}
\end{figure}


\textbf{On hardware, the geometric controller has fewer large excursions at the cost of higher velocity error.} PID and Mellinger demonstrate nearly identical position tracking, yet they diverge significantly in robustness and velocity regulation. The cascaded PID exhibits a \SI{4.20}{\percent} ($\pm$\SI{2.46}{\percent}) trajectory divergence rate, whereas the Mellinger controller reduces this to \SI{0.01}{\percent}, representing a significant improvement. This enhanced reliability stems from the Mellinger controller's $SE(3)$ formulation, which provides broader stability guarantees that prevent the aggressive attitude divergence occasionally observed with the PID. However, the Mellinger controller yields a velocity RMSE (\SI{0.89}{\meter\per\second}) that is $1.24\times$ higher than that of the PID (\SI{0.72}{\meter\per\second}). This trade-off is consistent with the geometric controller executing sharper attitude corrections, which inherently induce velocity transients on a nano-scale platform strictly bottlenecked by its \SI{0.6}{\newton} peak thrust authority. Heading errors remain comparable across both PID and Mellinger. Qualitative tracking performance and corresponding motor commands are visualized in Fig.~\ref{fig:task2_baselines_vs_gt} and Fig.~\ref{fig:task2_baselines_motors_overtime}, respectively.

\textbf{Offline-learned controllers are evaluated in simulation and cannot be compared directly with real-flight.} BC-MLP and BC-LSTM achieve position
RMSEs of \SI{0.15}{\meter} and \SI{0.18}{\meter} respectively, with zero
divergence, but these values reflect the idealized learned rollout plant rather
than physical flight readiness; the zero-variance confidence intervals across
all offline metrics confirm purely deterministic evaluation. MPPI diverges on
\SI{75.22}{\percent} of times (position RMSE \SI{1.32}{\meter}),
indicating that its sampling-based optimization is poorly matched to the narrow
actuation limits of sub-\SI{50}{\gram} platforms without domain-specific tuning.

\textbf{MPPI exhibits severe performance degradation within the nano-scale operational envelope.} MPPI yields a position RMSE of \SI{1.32}{\meter} and a \SI{75.22}{\percent} divergence rate. This indicates that the receding-horizon optimizer exploits inaccuracies of the learned forward model and that its horizon and sampling parameters require platform-specific tuning for the narrow actuation limits of sub-\SI{50}{\gram} vehicles. We report this configuration as a reference point rather than as a limit of sampling-based MPC on this platform.

\begin{table}[t]
\centering
\caption{Task~3: onboard EKF accuracy on trefoil trajectories, mean~$\pm$~standard deviation across all pooled runs ($N$ per speed regime).}
\label{tab:task3_trefoil}
\resizebox{\columnwidth}{!}{%
\begin{tabular}{@{}lcccccc@{}}
\toprule
\textbf{Speed} & \textbf{N} & \textbf{ATE RMSE (mm)} & \textbf{ATE Mean (mm)} & \textbf{RTE 1m (mm)} & \textbf{Vel RMSE (m/s)} & \textbf{Att RMSE (deg)} \\
\midrule
Slow & 11 & 21.1 $\pm$ 3.3 & 16.5 $\pm$ 3.2 & 28.2 $\pm$ 6.8 & 0.069 $\pm$ 0.015 & 2.12 $\pm$ 0.05 \\
Medium & 10 & 21.8 $\pm$ 3.0 & 17.2 $\pm$ 1.7 & 30.3 $\pm$ 2.9 & 0.067 $\pm$ 0.013 & 2.71 $\pm$ 1.43 \\
Fast & 7 & 3142.7 $\pm$ 3936.6 & 2774.4 $\pm$ 3487.9 & 384.2 $\pm$ 400.8 & 7.629 $\pm$ 8.992 & 29.67 $\pm$ 28.52 \\
\bottomrule
\end{tabular}}
\end{table}

\begin{figure}[htbp]
\centering
\includegraphics[width=1.0\columnwidth]{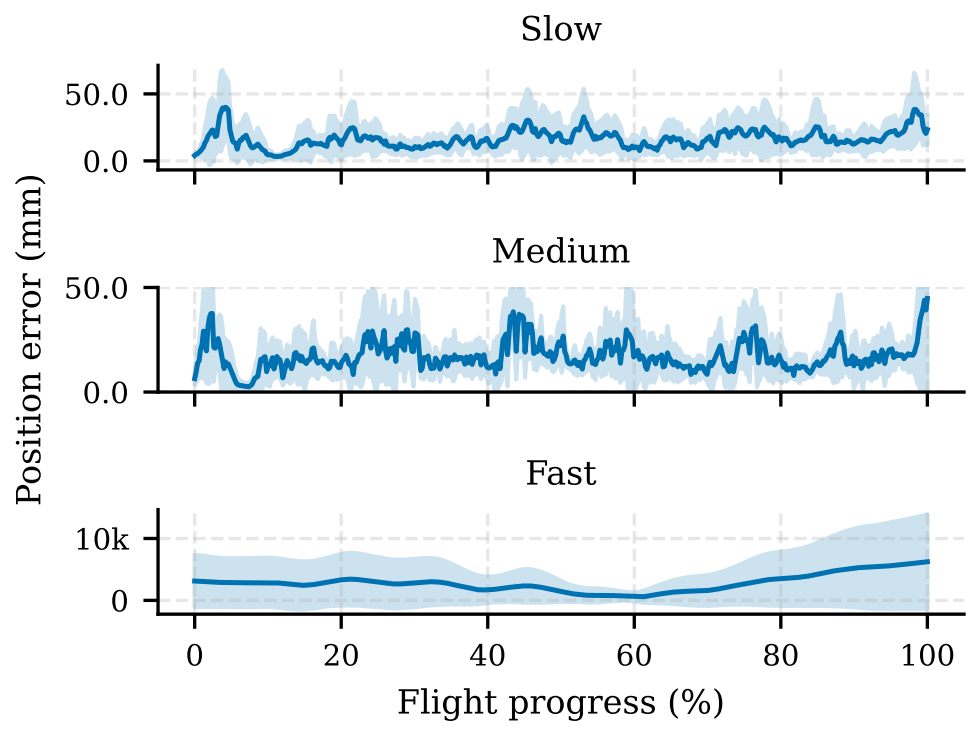}
\caption{Task~3: EKF position error (mean $\pm$ std, all runs pooled by speed) on trefoil trajectories. Error remains bounded at slow and medium speeds but diverges in the fast regime.}
\label{fig:task3_estimation}
\end{figure}

\subsection{Task 3: State Estimation}
\label{sec:results_estimation}

\subsubsection{Baselines}
\label{sec:estimation_baselines}

The \textbf{onboard EKF} runs at \SI{100}{\hertz} on the STM32F405 and fuses IMU data with the Vicon pose relayed through \texttt{send\_extpose()} to produce the 9-DoF state estimate used in flight. Because the EKF receives the Vicon pose as an external measurement, this task characterizes the consistency of the onboard fusion (its sensitivity to external-pose latency, packet loss, and process-model mismatch) rather than dead-reckoning accuracy. Runs from both real-flight controllers of Task~2 are pooled per speed regime. Metrics are ATE after Horn's $SE(3)$ alignment~\eqref{eq:ate}, RTE over \SI{1}{\meter} path-length windows, per-axis velocity residuals, and attitude RMSE in Euler angles. Because the estimator already operates in the Vicon frame, the recovered alignment is close to the identity, so the reported ATE is effectively a direct error in the common reference frame rather than an alignment-corrected one.

\subsubsection{Results and Analysis}
\label{sec:estimation_results}

Table~\ref{tab:task3_trefoil} reports EKF estimation accuracy on trefoil knot trajectories over three speed regimes, with all runs pooled (mean $\pm$ std across \num{28} runs). At slow and medium speeds, the onboard EKF maintains position tracking within \SI{22}{\milli\meter} ATE RMSE (\num{11} and \num{10} runs respectively), with attitude RMSE below \SI{3}{\degree} and velocity RMSE below \SI{0.07}{\meter\per\second}. At the fast regime (\SI{1.0}{\meter\per\second}), the EKF exhibits divergence (ATE $>$~\SI{3}{\meter} across \num{7} runs). The most plausible cause is loss or delay of external-pose packets at high angular rates on the shared radio link, from which the fixed-gain EKF does not recover within the flight; separating this effect from process-model mismatch is an open question that the released IMU, estimator, and ground-truth streams make possible to study. Fig.~\ref{fig:task3_estimation} shows the pooled position error over flight progress.

The three tasks are linked by the same platform and synchronized data: open-loop dynamics error (Task~1), closed-loop tracking (Task~2), and estimator accuracy (Task~3) all reflect the same hardware and trajectory set. NanoBench thus enables systematic comparison of methods across the autonomy stack on a platform for which no public benchmark previously existed.
\subsection{Limitations}
\label{sec:limitations}
Three limitations bound the present scope. First, synchronization estimates a single constant offset per flight; clock skew and drift are not modelled, the residual is bounded only by the \SI{1}{\milli\second} search grid rather than measured independently, and alignment quality degrades on trajectories with little rotational excitation, so the residual should not be assumed uniform across the dataset. Second, Task~3 supplies the Vicon pose to the estimator through \texttt{send\_extpose()} and evaluates against the same signal, which makes it a benchmark of Vicon-aided fusion consistency rather than of standalone onboard estimation; a dead-reckoning protocol requires re-running estimators offline from the released IMU stream. Third, Task~2's hardware and offline rows are distinct experiments on distinct plants, and the MPPI configuration is untuned, so neither supports a general claim about predictive control at this scale. All three are addressable with the released data and code, and we leave them as future work.
\section{Conclusion}
\label{sec:conclusion}
NanoBench introduces the first open-source, multi-task benchmark for nano-scale quadrotors, comprising 172 flight recordings on the commercially standard Crazyflie~2.1 in a Vicon motion capture arena. It is distinguished from prior aerial datasets by jointly providing synchronized motor PWM commands, onboard EKF and PID controller internals, and millimeter-accurate ground truth, signals that no existing benchmark exposes at the nano scale.

Baseline evaluations across three tasks expose platform-specific behaviors that larger-vehicle benchmarks cannot replicate. In Task~1, rigid-body dynamics achieve sub-millimeter one-step accuracy but diverge beyond a \SI{100}{\milli\second} horizon because of compounding thrust-model error, and a physics-plus-residual hybrid gives the lowest cumulative error. In Task~2, the default PID and Mellinger controllers reach similar tracking RMSE, but the geometric controller exhibits far fewer large excursions, while an untuned MPPI baseline fails on the learned plant. In Task~3, the onboard EKF maintains sub-\SI{22}{\milli\meter} ATE at slow and medium speeds but loses consistency in part of the fast-speed flights, a failure mode the released multi-stream logs allow future work to diagnose.


\bibliographystyle{IEEEtran}
\bibliography{IEEEexample}
\end{document}